\documentclass[sigconf]{acmart}
\AtBeginDocument{%
  \providecommand\BibTeX{{%
    \normalfont B\kern-0.5em{\scshape i\kern-0.25em b}\kern-0.8em\TeX}}}

\setcopyright{acmcopyright}
\copyrightyear{2019}
\acmYear{2019}
\acmDOI{preprint}

\acmConference[Submitted to KDD '19]{KDD '19: 2nd KDD Workshop on Anomaly Detection in Finance }{August, 2019}{USA}
\acmBooktitle{KDD '19: 2nd KDD Workshop on Anomaly Detection in Finance,August 04--08, 2019, Anchorage, Alaska - USA}
\acmPrice{15.00}
\acmISBN{978-1-4503-9999-9/18/06}

\begin{document}

%%
%% The "title" command has an optional parameter,
%% allowing the author to define a "short title" to be used in page headers.
\title{Infusing domain knowledge in AI-based "black box" models for better explainability with application in bankruptcy prediction}

%%
%% The "author" command and its associated commands are used to define
%% the authors and their affiliations.
%% Of note is the shared affiliation of the first two authors, and the
%% "authornote" and "authornotemark" commands
%% used to denote shared contribution to the research.
\author{Sheikh Rabiul Islam}
\email{sislam42@students.tntech.edu}
\affiliation{%
  \institution{Tennessee Tech University}
  \city{Cookeville}
  \state{TN}
  \postcode{38505}
}

\author{William Eberle}
\email{weberle@tntech.edu}
\affiliation{%
  \institution{Tennessee Tech University}
  \city{Cookeville}
  \state{TN}
  \postcode{38505}
}

\author{Sid Bundy}
\email{sbundy@tntech.edu}
\affiliation{%
  \institution{Tennessee Tech University}
  \city{Cookeville}
  \state{TN}
  \postcode{38505}
}

\author{Sheikh Khaled Ghafoor}
\email{sghafoor@tntech.edu}
\affiliation{%
  \institution{Tennessee Tech University}
  \city{Cookeville}
  \state{TN}
  \postcode{38505}
}

%%
%% By default, the full list of authors will be used in the page
%% headers. Often, this list is too long, and will overlap
%% other information printed in the page headers. This command allows
%% the author to define a more concise list
%% of authors' names for this purpose.
\renewcommand{\shortauthors}{Islam et al.}

%%
%% The abstract is a short summary of the work to be presented in the
%% article.

%% ***************************************** Abstract  *************************

\begin{abstract}
  Although ''black box'' models such as Artificial Neural Networks, Support Vector Machines, and Ensemble Approaches continue to show superior performance in many disciplines, their adoption in the sensitive disciplines (e.g., finance, healthcare) is questionable due to the lack of interpretability and explainability of the model. In fact, future adoption of ''black box'' models is difficult because of the recent rule of ''right of explanation'' by the European Union where a user can ask for an explanation behind an algorithmic decision, and the newly proposed bill by the US government, the ''Algorithmic Accountability Act'',  which would require companies to assess their machine learning systems for bias and discrimination and take corrective measures. Top Bankruptcy Prediction Models are A.I.-Based and are in need of better explainability\textendash{the extent to which the internal working mechanisms of an AI system can be explained in human terms}. Although explainable artificial intelligence is an emerging field of research, infusing domain knowledge for better explainability might be a possible solution. In this work, we demonstrate a way to collect and infuse domain knowledge into a "black box" model for bankruptcy prediction. Our understanding from the experiments reveals that infused domain knowledge makes the output from the black box model more interpretable and explainable. 
\end{abstract}

%%
%% The code below is generated by the tool at http://dl.acm.org/ccs.cfm.
%% Please copy and paste the code instead of the example below.
%%



%%
%% Keywords. The author(s) should pick words that accurately describe
%% the work being presented. Separate the keywords with commas.

%% A "teaser" image appears between the author and affiliation
%% information and the body of the document, and typically spans the
%% page.

%%
%% This command processes the author and affiliation and title
%% information and builds the first part of the formatted document.
\maketitle

%% ***************************************** Introduction  *************************

\section{Introduction}
Over the past few years, the field of Artificial Intelligence (AI) has gained enormous interest for its practical success in many application areas. Recent advances in machine learning and artificial intelligence have given rise to many complex and powerful models which are being adopted in many sophisticated areas such as medical, finance, and cyber-security. Some of the more notable models are Artificial Neural Networks (ANN), Genetic Algorithms (GA), Support Vector Machines (SVM), and Ensemble Approaches. Sometimes these models are called ''black box'' models.  Although the high complexity of the models'  non-linear functions come with good predictive power for black box models, they are limited by their \textit{explanation and interpretation} capabilities. Which is followed by a lack of trust in the model and the decisions it makes. 

In response to that issue, as a precaution to fight the unethical use of AI and biases in decision making, governments are trying to introduce and enforce new laws and regulations. For instance, the European Union implemented the rule of ''right of explanation'', where a user can ask for an explanation of an
algorithmic decision \cite{goodman2017european}. In addition, more recently, the US government has introduced a new bill called the ''Algorithmic Accountability Act'' \cite{algorithmic_accountability} which would require companies to assess their machine learning systems for bias and discrimination and take corrective measures. Should the bill pass, it will be enforced by the US Federal Trade Commission which is in charge of consumer protection and antitrust regulation.

Black box models are frequently used in the financial area, particularly towards bankruptcy prediction. In particular, the main focus of bankruptcy prediction is to predict the probability that the customer will be in default or bankrupt in the near future. The high rate of bankruptcy affects heavily the firm's owners, partners, society, and the overall economic condition of the country in general  \cite{alaka2018systematic}. According to literature reviews by Alaka et al. \cite{alaka2018systematic} and Bellovary et al. \cite{bellovary2007review}, six out of the top eight Bankruptcy Prediction Model (BPM) are Artificial Intelligence (AI) based. They also suggest that BPMs based on ''black box'' models such as ANN, GA, and SVM outperform all other models due to their capability of learning any non-linear function. Recently another black box model, ensemble approaches, where multiple models are combined for better results by correcting each other's error, shows promising performances. However, these ''black box'' models lack explainability (i.e., explaining the internal working mechanism to humans) and interpretability (i.e., a sense of what's happening), which raises ethical issues for domains like finance. A decision in the financial domain (e.g., credit approval, default prediction) needs to be more than a number\textemdash{} it needs to explain the reason behind the decision that makes sense to a human. Furthermore, when there are many explanatory variables, and some explanatory variables are complex, this further complicates the explainability. 

Research in Explainable Artificial Intelligence is an emerging field, seeing a resurgence after three decades of slowed progress since the work of Chandrasekaran et al. \cite{chandrasekaran1989explaining}, Swartout et al. \cite{swartout1993explanation}, and Buchanan et al. \cite{swartout1985rule}. In their work on Explainable Artificial Intelligence(XAI), Miller et al. \cite{miller2018explanation} argue that most of the work on XAI focuses on the researcher's intuition of what constitutes a good explanation. However, there exists a vast area of research in philosophy, psychology, and cognitive science on how people generate, select, evaluate, and represent explanations and associated cognitive biases and social expectations towards the explanation process. Therefore, the author emphasizes that, the research on explainable AI should incorporate study from these different domains. 

 According to Lipton et al.  \cite{lipton2016mythos} and \cite{on_explainable_artificial_intelligence}, interpretability has three different notions:
\begin{enumerate}
    \item interpretability in pre-modeling: finding and using simple, summarized, and relevant set of features from the domain; 
    \item interpretability in modeling: generating explanation along with the prediction to improve transparency;and
    \item interpretability in post modeling (a.k.a. post-hoc): understanding the dynamics between input and predicted output for an already trained/tested model.
\end{enumerate}
Unfortunately, the post-hoc notion of interpretability is not purely transparent and can be misleading, as it provides an explanation after the decision has been made. The algorithm can be optimized to placate subjective demand, and the explanation from it also can be misleading though it seems plausible (Lipton et al. 2016) \cite{lipton2016mythos}, \cite{on_explainable_artificial_intelligence}. Furthermore, from the literature review, we find that interpretability in pre-modeling is under-focused. Therefore, we particularly focus on the explainability of black box models \textit{using domain knowledge } which falls into interpretability in the pre-modeling stage. In this work, we take bankruptcy prediction as the context for our experiments.

  In our proposed approach, we replace hard to interpret features of a model with easily interpretable features (induced from domain knowledge) which allows the decision to be expressed in terms of an interpretable and concise set of features. We use a frequent pattern mining algorithm to find frequent feature sets used in different bankruptcy literature. Later, we relate the frequent feature set with the popular financial concept of credit to come up with a generalized feature set for the experiments, which ultimately allows us to infuse domain knowledge to increase the explainability and interpretability of "black box" models. To asses credit risk by human experts, the 5C's of credit is commonly used to analyze key factors: character (reputation of the borrower/firm), capital (leverage), capacity (volatility of the borrower's earnings), collateral (pledged asset) and cycle (macroeconomic) conditions \cite{angelini2008neural,5cs_of_credit}. The domain knowledge infused feature set gives us a generalized frequent feature set which is used for our experiments for better explainability. 

In summary, our contributions in this work are as follows: (1) we demonstrate a way to collect  and use domain knowledge from the literature; (2) we introduce a way to bring popular concepts (e.g., the 5 C's of credit) from literature to aid in interpretability and explainability; and (3) our experimental results show "black box" models can be better explainable with little or no compromise in performance when domain knowledge is infused.

We start with a background of related work (Section \ref{background}) followed by a description of our proposed approach and an overview of the dataset (Section \ref{methodology}) used in this work. In Section \ref{experiments}, we describe our experiments, followed by section \ref{sec:results} which contains results and a discussion of the experiments. We conclude with limitations and future work in section \ref{conclusion}.

%% ***************************************** Background  *************************

\section{background}\label{background}
Early research in explainable AI started with the preliminary work of  Chandrasekaran et al. \cite{chandrasekaran1989explaining}, Swartout et al. \cite{swartout1993explanation}, and Buchanan et al. \cite{swartout1985rule}. Recent advancements in AI, successful adoption in different applications, and awareness of ethical and bias issues necessitates have fueled recent research in  Explainable Artificial Intelligence (XAI). For instance, the DARPA division of the Department of Defense (DoD) is spending \$2 billion on its XAI program. They are developing a toolkit library consisting of machine learning and human-computer interface software for developing explainable AI systems that will be available for military and commercial use \cite{explainable_ai}.

Yang et al. \cite{yang2017explainable}, propose a method based on ''Bayesian Teaching'', where a subset of an example in used to train the model instead of the whole dataset. The subset of the example is chosen by domain experts that are most relevant to the problem. However, for this purpose, choosing the right subset of examples in the real world is challenging.

In sentiment analysis, the rationale for a prediction is important for understanding decisions. Lei et al. \cite{lei2016rationalizing} propose an approach that generates the rationale for a prediction. They demonstrate the approach with sentiment analysis from the text where a subset of text is selected as the rationale for the prediction. In addition, the selected text is concise and sufficient enough to act as a substitute for the original text, still capable of the correct prediction.  Although their approach outperforms available attention-based models, it is limited to only text analysis. 

Making a prediction that can be trusted is another challenge. Ribeiro et al. \cite{ribeiro2016should} propose a novel explanation technique capable of explaining the prediction of any classifier in an interpretable and faithful manner by learning an interpretable model locally around the prediction. Their concern is on two issues: (1) whether the user should trust the prediction of the model and act on that, and (2) whether the user should trust a model to behave reasonably well when deployed. In addition, they involve human judgment in their experiment to decide whether to trust the model or not.

Lundberg et al. \cite{lundberg2017unified} propose a unified approach called ''SHAP'' which unifies seven previous approaches: LIME \cite{ribeiro2016should}, DeepLIFT \cite{shrikumar2017learning}, Tree Interpreter \cite{on_tree_interpreter}, QII \cite{datta2016algorithmic}, Shapley sampling values \cite{vstrumbelj2014explaining}, Shapley regression values \cite{lipovetsky2001analysis}, and Layer-wise relevance propagation \cite{bach2015pixel} to make the explanation of prediction for any machine learning model. Both LIME \cite{ribeiro2016should} and SHAP \cite{lundberg2017unified} use a simplified input mapping, mapping the original input to a simplified set of input. However, none of the models incorporate domain knowledge. The following approach infuses domain knowledge into the experiment and works as a substitute for original complex features in order to generate a prediction which is explainable by itself.

%% ***************************************** Methodology  *************************

\section{Methodology}\label{methodology}
The proposed approach consists of two components:  a feature \textit{generalizer}, which gives a generalized frequent feature set with the help of domain knowledge, and an \textit{evaluator}, that produces and compares the results using the generalized feature set from the original feature set. 
\subsection{Feature Generalizer} \label{feature_generalizer_subsec}
\begin{figure}[h]
  \centering
  \includegraphics[width=\linewidth]{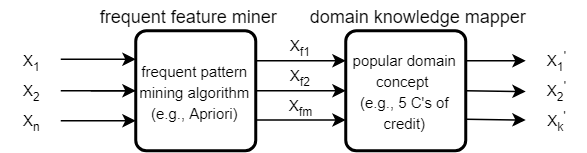}
  \caption{Feature generalizer}
    \label{fig:concept1_0}
\end{figure}

First, the frequent feature miner takes multiple different sets of features used in different bankruptcy prediction literature (see section 3.4) to discover the most frequent set of features (i.e., a frequent combination of features used in different literature) using a popular and classic frequent pattern mining algorithm called Apriori (Agrawal et al. \cite{agrawal1994fast}). In Figure \ref{fig:concept1_0}, the input to the algorithm is: 
\textit{X}\textsubscript{1}, \textit{X}\textsubscript{2},.... \textit{X}\textsubscript{n} \begin{math} \in \end{math} \textit{X}  where \textit{X} is the universal set of features used in mortgage bankruptcy prediction literature. The output is some frequent set of features with a specified \textit{support} and maximum count of features in the set:
\textit{X}\textsubscript{f1}, \textit{X}\textsubscript{f2},.....\textit{X}\textsubscript{m}  \begin{math} \in \end{math} \textit{X} where \textit{X} is the universal set of features as before, but here \textit{m} is much smaller than \textit{n}. 
Finally, the frequent set of features is fed into the domain knowledge mapper. In the domain knowledge mapper, a popular, easy to understand and interpret domain concept is introduced and mapped with the frequent feature set. For our case, we introduce the 5C\'s of credit which refers to capital, character, cash flow, conditions, and collateral. Based on the mapping, the  domain knowledge mapper outputs a generalized frequent feature set infused with domain knowledge.

\subsection{Evaluator} \label{evaluator_subsec}
\begin{figure}[h]
  \centering
  \includegraphics[width=\linewidth]{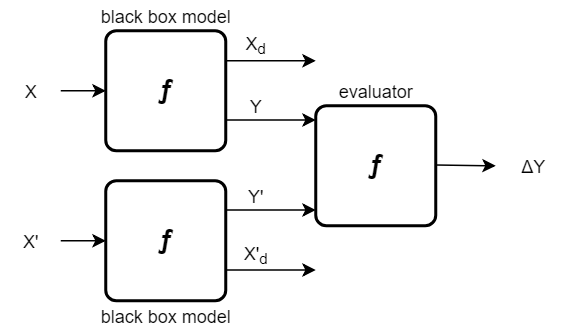}
  \caption{Evaluator}
  \label{fig:concept1_1}
\end{figure}

The task of the evaluator (Figure \ref{fig:concept1_1}) is to execute and compare the performance of two experiments: one using original features (\textit{X}) of the dataset and the other one using the generalized frequent feature sets (\textit{X}').  If the difference is within an allowable threshold, then the output from the latter experiment is deemed as final output, and the output is explained using the contribution from each of generalized, and more explainable and interpretable, frequent features.  

\subsection{Algorithms}
We use six different algorithms: one is for frequent pattern mining, and the remaining five are supervised ''black box'' models for predicting bankruptcy/default.
\subsubsection{Apriori}
The Apriori algorithm which was proposed by Agrawal et al. \cite{agrawal1994fast}, a classical algorithm in data mining for finding frequent patterns. For our case, when a set of features or explanatory variables found in a paper meets a user-specified support threshold, then that set of features can be treated as frequent feature sets. The support for a set of features X in a paper p\textsubscript{i} is defined as follows: Support (X) = (number of paper in which all features of X appear) / (total number of paper). For example, if the support threshold is set to .5 (i.e., 50\%), then the feature set {LTV, creditScore, interestRate, delinquencyStatus} is called a frequent feature set if and only if this set of features is found together at least 50\% of times among all the papers. Here is an intuition of the working mechanism of the Apriori algorithm. The Apriori algorithm iteratively finds frequent feature sets of a length starting from 1 to k, where k is the maximum number of features in any frequent feature set. The frequent feature sets must meet the minimum \textit{support} threshold of the algorithm. In addition, a subset of features from the frequent feature set must also be a frequent feature. For example,  if {LTV, creditScore, interestRate, delinquencyStatus} is a frequent feature set, then any of the features or any combination of features (e.g., {LTV} , {LTV, creditScore}) within this feature set is also a frequent feature set. In section \ref{experiments}, we will clarify this more.

\subsubsection{Artificial Neural Network (ANN)}
An Artificial Neural Network is a non-linear model, capable of mimicking human brain
functions. It consists of an input layer, multiple hidden layers, and the output layer. Each layer
consists of multiple neurons that help to learn the complex pattern, each subsequent layer learns more abstract concepts before it finally merges into the output layer. ANN was first
used in 1994 by Wilson and Sharda \cite{wilson1994bankruptcy} for bankruptcy prediction. 
In terms of different performance metrics, given enough data, ANN performs best for many problems due to its capability of learning any non-linear function.

\subsubsection{Support Vector Machine (SVM)}
The Support Vector Machine (SVM) was first introduced by Boser, Guyon, and Vapnik \cite{boser1992training} and has been used for many supervised classification tasks. The model learns an optimal
hyperplane that separates instances of different classes using a highly non-linear mapping of input
vectors in high dimensional feature space [Hooman, 2016]. SVM is listed as one of the top nonlinear algorithms for bankruptcy prediction in different literature surveys \cite{alaka2018systematic, bellovary2007review}. When the number of samples is too high (i.e., millions) then it is very
costly in terms of computation time. In that case, a non-linear algorithm like ANN can be a better choice as an ANN usually works well with large datasets.

\subsubsection{Random Forest (RF)}
A Random Forest is a tree-based ensemble technique developed by Breiman et al. \cite{breiman2001random}
for the supervised classification task. In RF, many trees are generated from the bootstrapped
subsamples (i.e., random sample drawn with replacement) of the training data. In each tree, the
splitting attribute is chosen from a smaller random subset of attributes of that tree (i.e., the chosen
split attribute that is the best among that random subset). This randomness helps to make trees less
correlated as correlated trees makes the same kinds of prediction errors and can overfit the model. This results in a forest of trees being generated, and the output from all the trees are averaged for the final
prediction. This averaging helps to reduce the variance from the model. Furthermore, RF can
work in a parallel computing environment as trees can be grown independently. According to \cite{fitzpatrick2016empirical}, RF has been used in different credit scoring and customer attrition applications.

\subsubsection{Extra Trees (ET)}
Extremely Randomized Trees or Extra Trees (ET) is also a tree-based ensemble
technique like RF and share a similar concept with RF. The only difference is in
the process of splitting attribute selection and determining the threshold (cutoff) value, both are
chosen in extremely random fashion \cite{islam2018mining}. As in RF, a random subset of
features are taken into consideration for the split selection but instead of choosing the most
discriminative cut off threshold, ET cut off thresholds are set to random
values. Thus, the best of these randomly chosen values is set as the threshold for the
splitting rule \cite{ensemble_methods}. As a result of multiple trees, the variance is reduced, compared to Decision Trees, however bias is introduced, as a subset of the whole feature set is chosen for
each tree. The ET which was proposed by Geurts et al.\cite{geurts2006extremely}, has continued its success by achieving the state of the art performance in some
anomaly/intrusion detection research \cite{islam2018efficient, islam2018credit, islam2018mining}.

\subsubsection{Gradient Boosting (GB)}
Friedman et al. \cite{friedman2001greedy}, generalized Adaboost to a Gradient Boosting algorithm that allows
a variety of loss function. Here the shortcoming of weak learners is identified using the gradient,
while in AdaBoost it is done through highly weighted data points. Gradient Boosting (GB) is a
classifier/regression model in the form of an ensemble of weak prediction models, such as  Decision Trees which are fitted with data initially. It also works sequentially like the 
AdaBoost algorithm, in that each subsequent model tries to minimize the loss function (i.e., Mean
Squared Error) by paying special focus on instances that were hard to get right in previous steps.

\subsection{Data}

We use two sources of data in this work:

\begin{enumerate}
    \item \textit{Explanatory variables dataset}: We went through the following 33 research papers  related to mortgage bankruptcy prediction: \cite{bhattacharya2019bayesian,greenwald2018mortgage,kvamme2018predicting,kim2018liquidity,fout2018credit,wu2018reducing,liu2018cure,karamon2017refinance,sirignano2016deep,
sousa2016new,chan2016determinants,
fitzpatrick2016empirical,sirignano2018risk,hooman2016statistical,tian2016unemployment,
moulton2016reducing,fang2016dynamics,antinolfi2016mortgage,berka2016using,bradley2015strategic,alfaro2012determinants,khandani2010consumer,
sousa2015links,sousa2015stress,anderson2014building,goodman2014look,foote2018mortgage,ghent2011recourse,elul2010triggers,
demiroglu2014state,low2015mortgage,gyourko2014reconciling,sorenson2015loan}
. We collected the explanatory variables used in each of these papers. We made the dataset available to the research community at here \cite{review_dataset}. Table \ref{tab:dataset} lists the features that appear four or more times in the literature.
    
    \item \textit{Freddie Mac single-family loan-level dataset}: The \textit{Freddie Mac} dataset \cite{freddie_mac} is the most frequently used dataset in the 33 previously mentioned research papers. It is also a publicly available dataset. For ensuring transparency, supporting the risk-sharing initiative, and building more accurate credit performance models, Freddie Mac, a government-sponsored enterprise, is making available loan-level credit performance data on fixed-rate mortgages that the company purchased or guaranteed. This is the source of data for the supervised algorithms used in this work. We took a stratified sample of the data to make sure the ratio of default vs non-default sample is same in both the original and the sample dataset. As the original dataset is an imbalanced dataset,  the sampled data contains 113,130 records, out of which only 198 of the records are defaults, giving us a highly imbalanced dataset with only .18\%  (<1\%) of target samples. In the anomaly detection problem, the class imbalance is not uncommon. In total there are 54 features in the dataset. We removed 24 unimportant features using feature ranking of the Random Forest algorithm, which gives us 30 features that we use for the experiments. Furthermore, we use 70\% of the data for training the models and kept 30\% of the data as a holdout set to test the model. We make sure the target class has the same ratio in both the training and test sets. 
\end{enumerate}

\begin{table*}
\caption{Frequent features found in different mortgage bankruptcy prediction literature with their appearance count and brief description}
\label{tab:dataset}
\centering
\begin{tabular}{llp{11cm}}
\toprule
Feature                      & Count &          Description                                                                                                          \\
\midrule  
creditScore                  & 26               & A number in between 300 and 850 that indicates the borrower's creditworthiness.                                                \\
LTV                          & 20               & Loan amount divided by the appraised value of the property.                                                                    \\
LTVoriginal                  & 13               & Original mortgage loan amount divided by the appraised value of the property on the note/purchase date.                        \\
creditScoreOriginal          & 12               & Credit score at loan origination time.                                                                                         \\
interestRateOriginal         & 10               & Original interest rate as indicated by the mortgage note.                                                                      \\
interestRateCurrent          & 9                & Active interest on the note.                                                                                                   \\
CLTVoriginal                 & 8                & Sum of all mortgage loans disclosed by the borrower divided by the apprised price of the mortgaged property on the note date.  \\
propertyState                & 8                & The territory of the property securing the mortgage.                                                                           \\
UPBoriginal                  & 8                & Unpaid principle balance on the note date.                                                                                     \\
postalCode                   & 7                & Denotes first three digits of five-digit postal code where the property is located.                                            \\
DebtToIncomeRatioOriginal    & 6                & Sum of monthly total debt payment divided by borrower's monthly income.                                                         \\
loanAge                      & 6                & Number of month passed since its origination.                                                                                  \\
CLTV                         & 6                & Sum of all mortgage loans disclosed by the borrower divided by the apprised price of the mortgaged property.                   \\
numberOfBorrowers            & 5                & Number of borrowers obligated to repay the loan.                                                                        \\
UPBactual                    & 5                & Unpaid principle balance as of latest month of payment.                                                                        \\
currentLoanDelinquencyStatus & 4                & Indicates the number of days the borrower is delinquent.                                                                       \\
numberOfUnits                & 4                & Indicates the number of unit in the property.\\
\bottomrule
\end{tabular}
\end{table*}

%% ***************************************** Experiments  *************************

\section{Experiments}\label{experiments}
First, in the feature generalizer (see section \ref{feature_generalizer_subsec}), for frequent feature mining, we use the Python-based library Mlxtend \cite{on_mlxtend}, which is actually an implementation of the Apriori algorithm. Second, in the evaluator (see section \ref{evaluator_subsec}), all supervised algorithms are implemented using the Python-based \textit{Scikit-learn} \cite{scikit-learn} library. In addition, we use Tensorflow \cite{tensorflow} for the Artificial Neural Network. We run all experiments on a laptop with 12GB of RAM and a core i7 processor. 

In other currently ongoing work, we investigated 33 research papers \cite{bhattacharya2019bayesian,greenwald2018mortgage,kvamme2018predicting,kim2018liquidity,fout2018credit,wu2018reducing,liu2018cure,karamon2017refinance,sirignano2016deep,
sousa2016new,chan2016determinants,
fitzpatrick2016empirical,sirignano2018risk,hooman2016statistical,tian2016unemployment,
moulton2016reducing,fang2016dynamics,antinolfi2016mortgage,berka2016using,bradley2015strategic,alfaro2012determinants,khandani2010consumer,
sousa2015links,sousa2015stress,anderson2014building,goodman2014look,foote2018mortgage,ghent2011recourse,elul2010triggers,
demiroglu2014state,low2015mortgage,gyourko2014reconciling,sorenson2015loan} related to mortgage default/bankruptcy prediction and collected all explanatory variables (i.e., features) \cite{review_dataset}. These collected features are the input data for the frequent feature mining algorithm.  The output is the frequent feature sets. The hyper-parameters for the frequent pattern mining algorithm (i.e., Apriori) are a minimum support threshold .05 and a maximum length 8. Here, \textit{support} for a set of feature(s) is the ratio of the number of research paper containing that feature(s) and the total number of the research paper. Furthermore,  maximum length refers to the maximum number of features that we want to see in any frequent feature set.
We brought 5 C's of credit as a concept from the domain and mapped the frequent features with the individual C's. We only keep the frequent feature sets that have at least one matching feature from each of the C's in the 5 C's of credit. Table \ref{tab:frequent_features_mapped} shows how we did the mapping and \ref{tab:frequent_features_matched} shows the mapped generalized frequent feature set. We wrote a python script \cite{project_code} to do this mapping. 

In the evaluator part, we use the Freddie Mac dataset for experimenting with the supervised algorithms ANN, SVM, RF, GB, and ET used in this work. We took a stratified sample of the data which contains 113,130 records, out of which 198 of the records are default giving us a highly imbalanced dataset with only .18\%  (<1\%) of target samples. Furthermore, we use 70\% of the data for training the models and kept 30\% of the data as a holdout set to test the model. We make sure the target class has the same ratio in both sets. We run the supervised algorithms in two different ways: \begin{enumerate}
    \item using original features: we use all 30 selected features given by the feature selection algorithm;
    \item using generalize frequent features set: we use each of the 25 generalized features sets separately for each algorithm. Each of the generalized feature sets consists of eight generalized feature based on the mapping from the domain knowledge (see \ref{tab:frequent_features_mapped} and \ref{tab:frequent_features_matched}). Out of the 25 runs for each algorithm with a different generalized feature set, we observe the performance and report the best performance with a corresponding generalized feature set in Section \ref{sec:results}.
\end{enumerate}

%% ***************************************** Results and Discussion  *************************

\section{Results and Discussion}{\label{sec:results}}
\begin{table*}[h!]
\centering
\caption{Some randomly chosen frequent feature set of length 8 and minimum support .05}
\label{tab:frequent_features}
\begin{tabular}{p{17cm}}
\toprule
Frequent Feature Set                                                                                                                                               \\
\midrule
\{UPBoriginal, LTV, LTVoriginal, creditScoreOriginal, interestRateCurrent, UPBactual, propertyState, creditScore\}                                                 \\
\{postalCode, interestRateCurrent, propertyType, loanTermOriginal, DebtToIncomeRatioOriginal, productType, propertyState, creditScore\}                            \\
\{postalCode, interestRateOriginal, interestRateCurrent, currentLoanDelinquencyStatus, CLTVoriginal, UPBactual, propertyState, creditScore\}                       \\
\{UPBoriginal, postalCode, interestRateCurrent, currentLoanDelinquencyStatus, CLTVoriginal, UPBactual, propertyState, creditScore\}                                \\
\{postalCode, interestRateOriginal, prepaymentPenaltyMortgageFlag, interestRateCurrent, productType, UPBoriginal, propertyState, creditScore\}                     \\
\{interestRateOriginal, interestRateCurrent, propertyType, loanTermOriginal, DebtToIncomeRatioOriginal, productType, UPBoriginal, prepaymentPenaltyMortgageFlag\} \\
\bottomrule
\end{tabular}
\end{table*}

The frequent pattern mining algorithm gives us a total of 4691 different combinations of feature sets. We discard feature sets that consist of less than eight features because frequent feature sets need to be big enough to cover at least one feature from each of the 5 C's.In addition, a few of the 5 C's are related to two or more features. By keeping combinations that consist of only eight features, we get 231 combinations of frequent feature sets. Table ~\ref{tab:frequent_features} exhibits few randomly chosen frequent feature sets of length 8.

\begin{table}[h]
\centering
\caption{Feature mapping with 5 C's of credit}
\label{tab:frequent_features_mapped}
\begin{tabular}{lp{6cm}}
\toprule
5 C's      & Mapped Feature from Frequent Feature Set                              \\
\midrule
Character  & creditScore, creditScoreOriginal, creditScoreCoborrower               \\
Capacity   & debtToIncomeRatioOriginal, currentDelinquencyStatus                   \\
Capital    & UPBactual, UPBoriginal                                                \\
Conditions & propertyState, interestRateCurrent, interestRateOriginal, postalCode  \\
Collateral & LTV, LTVoriginal, CLTV, CLTVoriginal                                  \\
\bottomrule
\end{tabular}
\end{table}

\begin{table*}[h!]
\centering
\caption{Frequent feature set that matches 5C's of credit}
\label{tab:frequent_features_matched}
\begin{tabular}{lp{16cm}}
\toprule
SL\# & Frequent Feature Set                                                                                                                                   \\
\midrule
1    & \{numberOfBorrowers, postalCode, interestRateOriginal, interestRateCurrent, currentLoanDelinquencyStatus, CLTVoriginal, UPBactual, creditScore\}       \\
2    & \{numberOfBorrowers, postalCode, interestRateOriginal, interestRateCurrent, currentLoanDelinquencyStatus, CLTVoriginal, UPBoriginal, creditScore\}     \\
3    & \{numberOfBorrowers, interestRateOriginal, interestRateCurrent, currentLoanDelinquencyStatus, CLTVoriginal, UPBactual, propertyState, creditScore\}    \\
4    & \{numberOfBorrowers, interestRateOriginal, interestRateCurrent, currentLoanDelinquencyStatus, CLTVoriginal, UPBoriginal, propertyState, creditScore\}  \\
5    & \{numberOfBorrowers, UPBoriginal, interestRateOriginal, interestRateCurrent, currentLoanDelinquencyStatus, CLTVoriginal, UPBactual, creditScore\}      \\
6    & \{postalCode, interestRateOriginal, interestRateCurrent, currentLoanDelinquencyStatus, CLTVoriginal, UPBactual, propertyState, creditScore\}       \\
\bottomrule
\end{tabular}
\end{table*}

Table ~\ref{tab:frequent_features_mapped} shows a mapping of the 5 C's to relevant features based on the information from \cite{angelini2008neural, 5cs_of_credit}. So far, we have 231 frequent feature sets irrespective of those containing at least one representative feature (based on mapping in ~\ref{tab:frequent_features_mapped}) from each C of the 5 C's of credit . We filter these frequent feature sets of length 8 by matching with the features mapping in Table \ref{tab:frequent_features_mapped} \textemdash{}all feature sets that don't contain at least one of the features from each category (each of the 5 C's) is discarded. This gives us 25 feature sets where each of the feature sets contains at least one of the features under each C of the 5 C's of credit. We call these 25 feature sets the \textit{generalized frequent feature sets}. Table ~\ref{tab:frequent_features_matched} shows some random generalized features sets.

\begin{figure}[h]
  \centering
  \includegraphics[width=\linewidth]{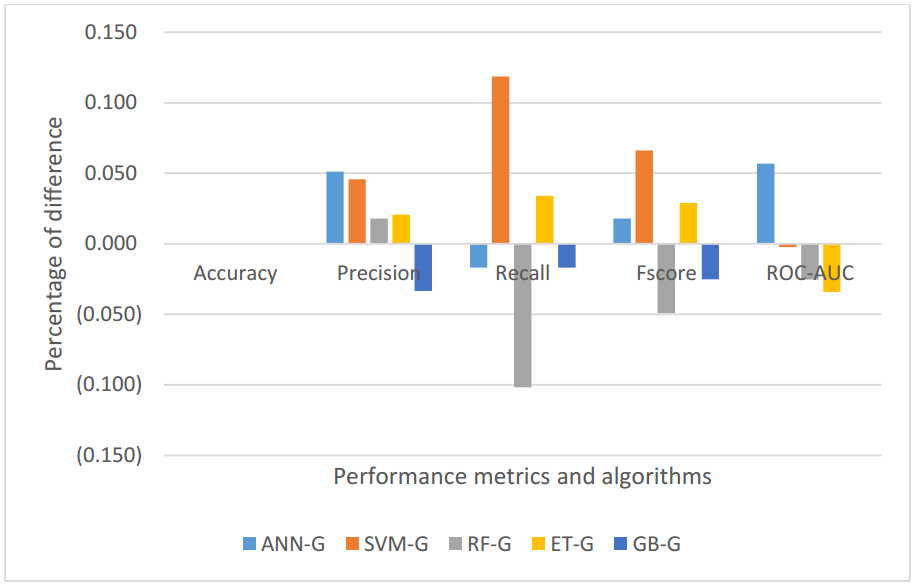}
  \caption{Dispersion in performance metrics for the case of using generalized frequent feature set}
  \label{fig:main_result_f}
\end{figure}

Table ~\ref{tab:mainresult} exhibits the performance comparison of different algorithms with or without using the generalized frequent feature set in terms of different performance metrics. An appended \textit{-G} after the algorithm name refers to when the algorithm is run using the generalized frequent feature set.  In addition, Figure \ref{fig:main_result_f} complements Table ~\ref{tab:mainresult} by providing the dispersion in performance metrics when using the generalized frequent feature set. Surprisingly, for all algorithms, there is no difference in accuracy in either of the cases when we use the generalized frequent features set or the original feature set. However, accuracy is not a good fit for our dataset to measure the performance due to a high imbalance in the data. The model can achieve a very high accuracy by classifying most of the samples as the majority class, which is misleading. Instead, recall, precision, fscore, and ROC-AUC are better measurements as it takes into account the misclassification errors (Type I error or false positives, Type II error or false negatives) that the model makes. In terms of precision, for all algorithms, performance drops slightly (in between 2 to 5\%) when using the generalized frequent feature set. In terms of recall and fscore, GB-G is the best and SVM-G is the worst. In terms of ROC-AUC, ANN-G is the worst and ET-G is the best. Overall, in terms of recall, precision, and fscore, the algorithm using the generalized frequent feature set performs better than when the same algorithms uses the original features of the dataset.

\begin{table}[h]
\caption{Comparison of recall, precision, and ROC-AUC between algorithms using original features and generalized frequent features}
\label{tab:mainresult}
\centering
\begin{tabular}{lllllll}
\toprule
Alg. & Acc. & Prec. & Rec.  & F.  & AUC & Time     \\
\midrule
ANN       & 0.999    & 0.845     & 0.831   & 0.838   & 0.980   & 458.852  \\
ANN-G     & 0.999    & 0.794     & 0.847   & 0.820   & \textbf{\textit{0.924}}   & 23.259   \\
          & 0.000    & 0.051     & (0.017) & 0.018   & 0.057   & 435.593  \\
SVM       & \textbf{\textit{0.997}}    & 0.356     & 0.881   & 0.507   & \textbf{0.995}   & \textbf{955.312}  \\
SVM-G     & \textbf{\textit{0.997}}    & \textbf{\textit{0.310}}     & \textbf{\textit{0.763}}   & \textbf{\textit{0.441}}   & 0.997   & 491.200  \\
          & 0.000    & 0.046     & 0.119   & 0.066   & (0.002) & 464.112  \\
RF        & \textbf{1.000}    & \textbf{1.000}     & 0.831   & 0.907   & 0.958   & 12.371   \\
RF-G      & \textbf{1.000}    & 0.982     & 0.932   & 0.957   & 0.983   & \textbf{1.948}    \\
          & 0.000    & 0.018     & (0.102) & (0.049) & (0.025) & 10.423   \\
ET        & \textbf{1.000}    & \textbf{1.000}     & 0.831   & 0.907   & 0.966   & 219.501  \\
ET-G      & \textbf{1.000}    & 0.979     & 0.797   & 0.879   & \textbf{1.000}   & 5.576    \\
          & 0.000    & 0.021     & 0.034   & 0.029   & (0.034) & 213.925  \\
GB        & \textbf{1.000}    & 0.932     & 0.932   & 0.932   & 0.999   & 625.230  \\
GB-G      & \textbf{1.000}    & 0.966     & \textbf{0.949}   & \textbf{0.957}   & 0.999   & 373.387  \\
          & 0.000    & (0.033)   & (0.017) & (0.025) & 0.000   & 251.843 \\
\bottomrule
\end{tabular}
\end{table}

Furthermore, from Figure 4, we can also see that, for the most important performance metric recall, we are getting better performance (ranging from 2-10\%) for ANN-G, RF-G, and GB-G. In terms of any performance metrics, there is at least one algorithm that performs better or equal using the generalized frequent feature. We need to choose the right algorithm based on the class distribution in the data as performance metric response differs based upon the distribution of the class in the data. Furthermore, when we use the generalized feature sets, we run the algorithm on all 25 generalized feature sets to discover the best feature sets. We found that using frequent feature set \# 5 (see Table \ref{tab:frequent_features_matched}), out of the 25 generalized feature sets, for algorithms RF-G, ET-G, and GB-G we are able to achieve the best result based on performance metric \textit{recall}.  For ANN-G and SVM-G, pattern 3 and 6 worked better accordingly. Therefore, this helps to choose the best generalized frequent feature set for a particular algorithm among many generalized frequent feature sets. 

We only tested with the \textit{Freddie Mac} dataset and there is a chance that the original features (even after excluding unnecessary features using the feature selection technique) still overfit the model, which leads to a better or equal result for all performance metrics in our case. Validating the result with multiple datasets is part of the future direction of this work. Furthermore, overall, all algorithms using a generalized frequent feature set takes less execution time compared to their counterparts due to the much fewer number of features. For a few algorithms (e.g., ANN), a fewer number of features decreases the computation time. 

Overall, even though infusing domain knowledge might lead to some compromise in performance, clearly it ensures better explainability and interpretability as the output is made from a concise and familiar set of features from the domain. Our success so far is in the generation of the output using an intuitive set of features. Our further concern is to show the result in an interpretable way. One way is by expressing the output as a percentage of the total risk, and the segregation of the output value is the percentage that each of the generalized frequent features is liable. We can express the total risk probability with the following formula:
\begin{equation} \label{eq:formula1}
 P(D) = \sum_{g=0}^{G} contribution (g)
\end{equation}

where g is the generalized frequent feature. Instead of using  contributions from generalized frequent features, we can also express the output in terms of the contribution from each element of the domain concept. This might improve the interpretability a little bit at the expense of losing some details. 

The correct way to come up with the breakdown of contribution from each feature for a particular prediction contribution is challenging. The naive way to formulate this can be by using the importance or permutation importance of the features. However, the importance of the feature is usually calculated based on a set of data (e.g., training set) and can be achieved directly from feature importance methods in most supervised algorithms. However, in case of sample wise feature importance this is not for straight forward. Moreover, the test sample might not be a good representative of the training set.  Other work such as LIME \cite{ribeiro2016should}, Tree Interpreter \cite{on_tree_interpreter}, SHAP \cite{lundberg2017unified}, and ELI5 \cite{on_eli5}, can discover the contributions of features in the prediction. However, each of the available techniques/tools come with some limitations: some are applicable to only text or images individually, and some are applicable to only a class of algorithms (e.g., tree based approaches, neural networks). Most of these approaches try to find out how the prediction deviates from the base/average scenario. Lime \cite{ribeiro2016should} tries to generate an explanation by locally (i.e., using local behavior) approximating the model with an interpretable model (e.g., decision trees, linear model). However, it is limited by the use of a linear model to approximate local behavior.Furthermore, SHAP unifies previous approaches including LIME by borrowing features from those. While SHAP comes with theoretical guarantees about consistency and local accuracy from game theory, in the case of black box kernel SHAP, it needs to run many evaluations of the original model to estimate a single vector of feature importance \cite{shap_vs_lime}. ELI5 also uses the LIME algorithm internally for explanations, however, the model is not truly agnostic, mostly limited to tree-based and other parametric or linear models. Tree Interpreter is limited to only tree-based approaches (e.g., Random Forest, Decision Trees). Our future work includes finding an optimal solution for sample-wise feature contribution in the prediction and express the sample-wise output according to the formula \ref{eq:formula1}.

\section{Conclusions and future work}\label{conclusion}
Future adoption of "black box" models is in an inauspicious position due to the lack of explainability. Governments of different countries have started to introduce laws to ensure accountability, right of explanation, and eliminating bias/discrimination in  decisions. Sophisticated areas such as finance, security,  and healthcare are in need of better explanations of their  "black box" models. In this work, we demonstrated a way to collect and use domain knowledge from the literature. We also introduced a way to bring and infuse popular concepts (e.g., 5 C's of credit) from the literature that aide in better interpretability and explainability. Our experimental results show that "black box" models can be better explainable without much compromise in performance when domain knowledge is infused.

Experimenting with our proposed approach on multiple data sets will help us validate its versatility. Moreover, finding an optimal solution to segregate the contribution of each participating feature (sample wise) will aide in better explainability of sample wise output and better run-time of the model. In addition, incorporating other concepts as domain knowledge will verify its generality, making this approach transferable to other domains like cyber-security and healthcare. For instance, in cyber-security, better explanation would aide in the   understanding of different attack scenarios and safeguarding the model from adversarial attacks. 

%%
%% The acknowledgments section is defined using the "acks" environment
%% (and NOT an unnumbered section). This ensures the proper
%% identification of the section in the article metadata, and the
%% consistent spelling of the heading.
\begin{acks}
Thanks to Tennessee  Tech  Cyber-security  Education, Research and Outreach  Center  (CEROC) for funding this research.
\end{acks}

%%
%% The next two lines define the bibliography style to be used, and
%% the bibliography file.
\bibliographystyle{ACM-Reference-Format}
\bibliography{sample-base}

%%
%% If your work has an appendix, this is the place to put it.
\appendix

\end{document}